\documentclass{article}
\usepackage{fancyvrb}
\usepackage{xcolor}
\usepackage{graphicx}
\definecolor{mybg}{rgb}{1,1,0.8}
\usepackage{xcolor} 
\usepackage{tcolorbox}
\tcbuselibrary{listings, breakable, skins}

\newtcblisting{customlisting}[1][]{
    breakable,
    listing only,
    colback=lightgray!20,
    colframe=black,
    enhanced, 
    listing options={
        basicstyle=\ttfamily\small,
        breaklines=true,
        breakatwhitespace=false,
        columns=fullflexible
    },
    title={#1}
}
\usepackage[preprint]{neurips_2025}

\usepackage[utf8]{inputenc} 
\usepackage[T1]{fontenc}    
\usepackage{hyperref}       
\usepackage{url}            
\usepackage{booktabs}       
\usepackage{amsfonts}       
\usepackage{nicefrac}       
\usepackage{microtype}      
\usepackage{xcolor}         
\usepackage{amsmath}
\usepackage{algorithm}
\usepackage{algpseudocode}
\usepackage{pgfplots}
\pgfplotsset{compat=1.17}

\title{Towards Reliable ML Feature Engineering via Planning in Constrained-Topology of LLM Agents}

\author{%
 Himanshu Thakur\thanks{Work done while working at JPMorgan Chase.}\\
 Meta\\
 Menlo Park, CA\\
  \texttt{
scihimanshu10@gmail.com} \\
  \And
  Anusha Kamath \\
  JPMorgan Chase \& Co.\\
  Palo Alto, CA \\
  \texttt{anusha.kamath@chase.com} \\
  \AND
  Anurag Muthyala\\
  JPMorgan Chase \& Co.\\
  Bangalore, India \\
  \texttt{anurag.muthyala@chase.com} \\
  \And
  Dhwani Sanmukhani \\
  JPMorgan Chase \& Co.\\
  Palo Alto, CA \\
  \texttt{dhwani.sanmukhani@chase.com} \\
  \And
  Smruthi Mukund \\
  JPMorgan Chase \& Co.\\
  Palo Alto, CA \\
  \texttt{smruti.mukund@chase.com} \\
  \And
  Jay Katukuri \\
  JPMorgan Chase \& Co.\\
  Palo Alto, CA \\
  \texttt{jay.katukuri@chase.com} \\
}

\begin{document}

\maketitle

\begin{abstract}  

Recent advances in code generation models have unlocked unprecedented opportunities for automating feature engineering, yet their adoption in real-world ML teams remains constrained by critical challenges: (i) the scarcity of datasets capturing the iterative and complex coding processes of production-level feature engineering; (ii) limited integration and personalization of widely used coding agents, such as CoPilot and Devin, with a team’s unique tools, codebases, workflows, and practices; and (iii) suboptimal human-AI collaboration due to poorly timed or insufficient feedback. We address these challenges with a planner-guided, constrained-topology multi-agent framework that generates code for repositories in a multi-step fashion. The LLM-powered planner leverages a team’s environment, represented as a graph, to orchestrate calls to available agents, generate context-aware prompts, and use downstream failures to retroactively correct upstream artifacts. It can request human intervention at critical steps, ensuring generated code is reliable, maintainable, and aligned with team expectations. On a novel in-house dataset, our approach achieves 38\% and 150\% improvement in the evaluation metric over manually crafted and unplanned workflows respectively. In practice, when building features for recommendation models serving over 120 million users, our approach has delivered real-world impact by reducing feature engineering cycles from three weeks to a single day.

\end{abstract}
\vspace{-1em}
\section{Introduction}

In the rapidly evolving field of AI-driven code generation, integrating autonomous agents into real-world engineering teams raises significant challenges in reliability, adaptability, and efficiency. Tools such as Devin, Claude Code, and Cursor promise to automate project creation but still require substantial human oversight and struggle to adapt to team-specific workflows, codebases, and in-house tools \cite{gu2025challengespathsaisoftware}. These challenges are especially pronounced in feature engineering for machine learning, where agents must reason about unit tests, ensure logical correctness rather than mere syntax, and handle delayed outputs from distributed runtimes such as PySpark. Agents also need knowledge of data distributions, feature completeness checks, and the ability to proactively handle changes in underlying datasets, increasing automation complexity \cite{zhang2025dynamicadaptivefeaturegeneration}.

Recent advances in large language models (LLMs) have enabled progress in code generation ~\cite{huynh2025largelanguagemodelscode,trirat2025automlagentmultiagentllmframework,jin2025towards}, but single-shot approaches struggle in feature engineering: they treat artifacts independently, lack awareness of interdependencies, and cannot exploit delayed validation signals. Many agents cannot measure correctness locally, so outputs are often validated only via the full episode or downstream agents, making  detection difficult. Open-ended multi-agent frameworks such as AutoML-GPT~\cite{zhang2023automlgptautomaticmachinelearning}, OpenDevin~\cite{wang2025openhandsopenplatformai}, LangGraph orchestrators, and others~\cite{applis2025unified,soni2025coding} provide flexibility but suffer from unbounded exploration, poor coordination, and limited correctness guarantees. Conversely, master-slave architectures enforce rigid flows, limiting adaptivity and preventing effective upstream error recovery~\cite{sun2025multiagentapplicationofficecollaboration}. Real-world environments impose additional constraints: many agents are fixed and can only be modified via prompting, necessitating frequent manual tuning, while team workflows evolve rapidly, making updates harder~\cite{wang2024agentworkflowmemory, zou2024cooperative, muhoberac2025state, wang2025mirix, xu2025mem, li2023tradinggpt}.

\begin{figure*}[t]
    \centering
    \includegraphics[width=0.95\textwidth]{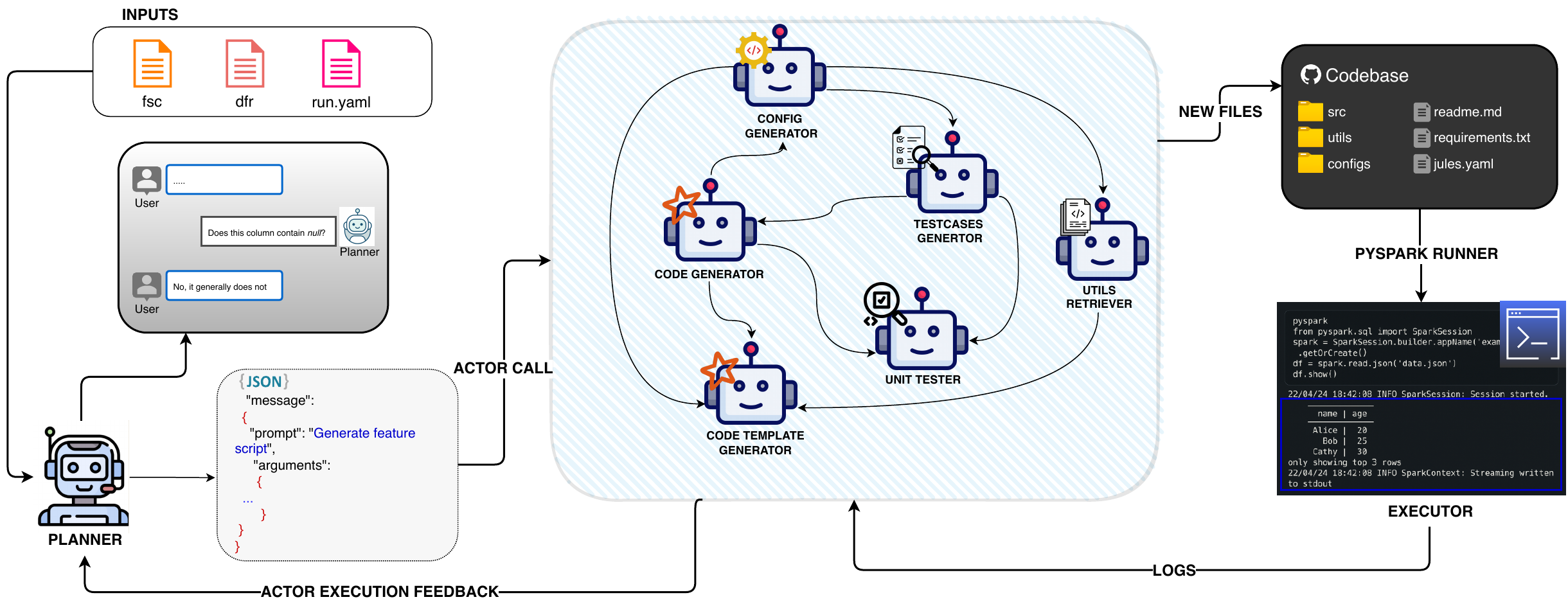}
    \vspace{-0.3cm}
    \caption{Overview of our approach, illustrating interactions between the planner, actors, and executor. Actors and their dependencies match the actual the development environment $E$ used in experiments.}
    \label{fig:mainfig}
    \vspace{-0.2cm}
\end{figure*}

To address these limitations, we propose a planner-guided, constrained-topology multi-agent framework for reliable, multi-step feature engineering and repo-level code generation. The framework models a team’s environment and codebase as a constrained topology, allowing an LLM-based planner to dynamically orchestrate agents, generate context-aware prompts, and leverage downstream failures—such as Spark errors or failing tests—to retroactively correct upstream outputs. It supports human-in-the-loop collaboration and extensibility: teams can define custom agents and topologies without retraining, and the planner can consult humans in ambiguous cases, reducing cognitive overhead. By prioritizing validated, human-written code and enforcing structured agent interactions, our approach improves reliability, maintainability, and human understanding of generated artifacts.

Our main contributions can be summarized as follows:

\begin{enumerate}

\item We propose a planner-guided, constrained-topology framework of LLM-driven agents for reliable, production-ready, repository-level code generation, enabling adaptive agent selection and optimized multi-turn execution.
\item We demonstrate that an LLM-based planner coordinating existing agents achieves significant gains in complex feature engineering tasks, while seamlessly integrating human-in-the-loop interventions to ensure \textit{reliable}, maintainable, and workflow-aligned pipelines.
\item We devise a first-of-its-kind, PySpark-based, multi-turn, repository-level benchmarking dataset that faithfully mirrors real-world production ML feature engineering pipelines.
\end{enumerate}
\vspace{-1em}

\section{Methodology}
\vspace{-0.5em}
In this section, we present an overview of our multi-turn, multi-agent coding framework within a structured environment, detailing the environment, planner, executor, actors, and memory along with their key design considerations.

\textbf{Environment:} The environment $E$ is a constrained-topology directed graph $G = (V, \hat{E})$, where nodes $v \in V$ correspond to workflow steps linked to actors $\mathcal{A}$, and edges $\hat{E}$ define valid execution paths (see more in Appendix \ref{tab:actor-topology}). This topology enforces dependencies while allowing the planner to dynamically select the optimal actor sequence within these constraints. Each task is described using three inputs: (i) \texttt{FSC} (Feature Specification Config), a YAML file specifying target features, base columns and their datasets, and computation logic; (ii) \texttt{DFR} (DataFrame Registry), a YAML file describing base datasets and dependencies; and (iii) a run file specifying the codebase README, reusable utilities, and working repository. The README is assumed to contain guides and practices needed to contribute to the codebase.

\textbf{Planner: }
The planner $P$ is an LLM-powered agent responsible for identifying the optimal path through the environment graph $G$ to complete a task. It selects the next actor $A_i$ at each step $t$ based on the current state $s_t$ and the constrained topology of $G$, which encodes permissible transitions between actors. The state $s_t$ includes the status of all actors, previous inputs and outputs, encountered errors and proposed and applied fixes, stored in short-term memory $M_{s_t}$. Using this information, the planner generates context-aware instructions, decides which actor to call next, and determines when human confirmation or intervention is required—for example, to verify an actor without explicit success criteria or to resolve repeated errors.

\textbf{Actors: } Actors, denoted as $\mathcal{A} = {A_1, A_2, \ldots, A_m}$, are the nodes of the graph $G$ and can be tools or LLM agents executing specific workflow sub-tasks. Each actor $A_i$ is defined by its input/output (I/O) signature and a description of the sub-task it performs. This enables the executor to provide the correct inputs and the planner to generate optimal instructions. The interaction is formalized as $A_i(x) = f(x, \theta_i)$, where $x$ is the relevant input context and $\theta_i$ is the fixed parameter set supplied by the executor. To handle the complexities of evaluating workflow stages—especially when an actor’s success depends on others—we adopt a flexible approach allowing actors to operate with or without strict success criteria. Each actor has a loose boolean success function $S_i(x)$, where $S_i(x) = \text{True}$ indicates success, as summarized in Table~\ref{tab:actor_descriptions}. End-stage actors, such as the code generator and test case coder, are assigned stricter criteria to ensure reliable completion. To improve robustness, actors can retry tasks using access to all previous errors and execution traces. Inspired by ReACT \cite{yao2023reactsynergizingreasoningacting}, actors generate the reason for failure, propose a fix, and produce corrected output. Each actor can autonomously decide whether to retry or request planner intervention by outputting the keyword \texttt{TERMINATE} in the proposed fix tag.

\textbf{Planner-Guided Actor Execution: } Now, we describe our proposed approach. Given a task $T$ in environment $E$, represented as a graph $G$, execution begins with a \texttt{FSC} file specifying target features, the \texttt{DFR} containing sample datasets and schemas, and locations of the codebase README and reusable utilities. The planner $P$ coordinates the entire multi-turn process, determining at each step $t$: (i) which actor to invoke, (ii) whether human intervention is required, and (iii) the textual prompt to provide to the actor. Once an actor is selected, the planner delegates execution to the \textit{executor}, which serves as the intermediary between the planner and the actors by preparing fixed inputs, attaching planner instructions, and retrieving outputs. For example, if the code generation actor fails, the planner analyzes reported errors to identify whether the failure stems from the code generator itself or from an upstream actor, such as the config generator, that provided incorrect inputs. Intermediate results are stored in short-term memory $M_{s_t}$, allowing the planner to refine its strategy based on past successes and failures. Actors may retry up to $K=5$ times, and persistent failures trigger the planner. The process concludes when all invoked actors satisfy their success criteria, and all generated artifacts—such as configuration files, PySpark scripts, and unit tests—are seamlessly integrated into the working repository via a pull request.

\vspace{-1em}
\section{Experiments and Results}
\vspace{-0.5em}
In this section, we provide a comprehensive overview of the dataset utilized, along with detailed descriptions of the experiments conducted. We outline the specific metrics captured during these experiments and present the results that demonstrate the effectiveness of our proposed approach. Furthermore, we conduct a comparative analysis between our approach and two distinct baseline methodologies, highlighting the necessity and advantages of our approach within the proposed setting.

\textbf{Dataset:} We introduce a newly developed benchmark dataset that closely mirrors real-world industrial environments for machine learning featurization, specifically leveraging PySpark for large-scale data processing. Unlike community datasets such as SWE-bench \cite{jimenez2024swebenchlanguagemodelsresolve} or MLE-bench \cite{chan2025mlebenchevaluatingmachinelearning}, which primarily utilize Pandas, our dataset is tailored to the complexities of production-scale workflows. It comprises 10 tasks that involve generating PySpark scripts, unit tests, and configuration files, each focused on constructing features for user-offer recommendation pipelines in e-commerce environments. These tasks vary in complexity and are designed to cover a representative subset of common featurization actions. Additionally, a separate held-out (0th) task is used exclusively for developing our proposed approach and is not included in benchmarking. Each task requires the creation of new feature scripts that perform a combination of operations, as summarized in Table~\ref{tab:data_processing_operations}. The dataset’s unique repository structure and coding conventions ensure adaptability to diverse team-specific formats.

\textbf{Experiments:} We compare our approach against two baselines under identical experimental conditions. Both baselines exclude planner-provided input prompts; only the planner differs. For benchmarking purposes, human input requests by the planner are met with a default response indicating that help is not available. \textbf{Sequential Agent Selection} follows a fixed, deterministic workflow, invoking actors in a predefined order. \textbf{Graph-Constrained Random Actor Selection} chooses actors uniformly at random from eligible next steps along valid graph paths, continuing for up to $N$ iterations or until task completion. Each approach is run three times per task, and we report the mean performance. We define a total of six actors, with detailed descriptions provided (see Appendix~\ref{actors_overview}). Our primary metric is \textit{pass@3}, the fraction of successful runs out of three. We also report task- and actor-level metrics (see Appendix \ref{sec:appendix_actors}). For all experiments, we use Claude 3.7 Sonnet as our LLM with a temperature of 0.1 and a maximum of 8192 tokens. 

\textbf{Results:} Table \ref{tab:main_results} compares the mean \textit{pass@3} scores of our approach, \textbf{Planner-Guided Actor Execution}, against two baselines. \textbf{Sequential Actor Selection}, which follows a fixed actor workflow, achieves 0.600, while \textbf{Graph-Constrained Random Actor Selection} reaches 0.333. Our planner-guided framework outperforms both with a mean pass@3 of 0.833, highlighting the advantage of dynamic, planner-driven actor selection for complex, multi-step tasks. The standard deviations, reported in parentheses, indicate consistent improvements across runs and tasks. Task-level metrics, including planner steps and actor failure rates, are provided in Table \ref{tab:task_metrics_summary}.

\begin{table}[ht]
\centering
\caption{Pass@3 metric computed across runs and averaged across tasks, reported as mean (stddev).}
\label{tab:main_results}
\begin{tabular}{lc}
\toprule
\textbf{Algorithm} & \textbf{pass@3} \\
\midrule
Sequential Actor Selection & 0.600 (0.490) \\
Graph-Constrained Random Actor Selection & 0.333 (0.471) \\
Planner-Guided Actor Execution (Ours) & \textbf{0.833 (0.373)} \\
\bottomrule
\end{tabular}
\end{table}
\vspace{-1em}
\section{Related Works}
\vspace{-0.5em}
Recent work has explored LLM-based agents for software development, which can be broadly grouped intogo/ human-AI collaboration and multi-turn code generation. In human-AI collaboration, \cite{hula, pasuksmit2025human, retzlaff2024human} propose a planner-coding-human agent system that emphasizes plan confirmation with humans. While effective for linear workflows, frequent human intervention slows execution and does not handle revisiting steps when success criteria are unclear. Extensions such as \cite{feng2024largelanguagemodelbasedhumanagent} fine-tune models to optimize human intervention, but this can be costly and less accessible for smaller teams. In multi-turn code generation, LLM-based agents \cite{xi2023risepotentiallargelanguage,Wang_2024,guo2024largelanguagemodelbased} integrate perception, memory, decision-making, and action to autonomously generate code. Multi-agent systems \cite{liu2025sewselfevolvingagenticworkflows, wang2025evoagentx, zhang2025evoflow} use evolutionary prompting to generate and validate code, while \cite{cafe, hollmann2023caafe} iteratively produces semantically meaningful features for tabular datasets. Multi-turn approaches \cite{zeng2025reinforcingmultiturnreasoningllm} treat code generation as a sequence of steps, optimizing over output formats, tool use, and failures. Despite these advances, existing methods often rely on zero-shot prompting or limited context, limiting their ability to handle multi-step, repo-level code generation in complex workflows.
\section{Conclusions}
\vspace{-0.5em}
We presented a multi-agent system for reliable, multi-step ML feature engineering and repository-level code generation, with a central planner orchestrating agent interactions. By dynamically selecting actors, managing workflow dependencies, and leveraging downstream feedback, the planner enables robust, high-quality code generation, outperforming manually crafted and unplanned workflows. Experiments on realistic PySpark pipelines demonstrate substantial gains over baselines, quantitatively highlighting the planner’s impact on task success. This work underscores the effectiveness of structured, planner-guided orchestration for scalable, adaptable, and autonomous code generation in real-world ML workflows.
\vspace{-0.5em}
\section{Limitations and Future Works}
\vspace{-0.5em}

While our planner-guided framework demonstrates strong performance, it has several limitations. It relies on fixed prompting and downstream validation, which can delay error detection and reduce adaptability to diverse coding standards and workflows. Our current task dataset is relatively small, limiting the diversity of scenarios; expanding it could better capture real-world feature engineering challenges. Future directions include fine-tuning the planner LLM to improve plan generation, incorporating long-term memory for better context retention, and exploring collaborative multi-agent strategies to handle more complex and interdependent code generation tasks. Together, these enhancements could increase robustness, scalability, and overall effectiveness in automating production-level ML feature engineering pipelines. 

\vspace{-0.5em}
\section{Disclaimer}
This paper was prepared for informational purposes by the Artificial Intelligence Research group of JPMorgan Chase \& Co. and its affiliates (“JP Morgan’’) and is not a product of the Research
Department of JP Morgan. JP Morgan makes no representation and warranty whatsoever
and disclaims all liability, for the completeness, accuracy, or reliability of the information
contained herein. This document is not intended as investment research or investment
advice, or a recommendation, offer or solicitation for the purchase or sale of any security,
financial instrument, financial product, or service, or to be used in any way for evaluating the
merits of participating in any transaction, and shall not constitute a solicitation under any
jurisdiction or to any person, if such solicitation under such jurisdiction or to such person
would be unlawful.
© 2024 JPMorgan Chase \& Co. All rights reserved 
\vspace{-0.5em}

\bibliographystyle{plain}
\bibliography{references}

\appendix

\section{Actors Overview}
\label{actors_overview}

Actors in our multi-agent framework are specialized agents or tools executing distinct sub-tasks in the feature engineering workflow. Each actor has a defined input/output signature and clear responsibilities, enabling modular orchestration by the planner and executor.

Table~\ref{tab:actor_descriptions} summarizes the actors for the featurization task along with their success criteria, ensuring outputs meet production-level quality and reliability standards.

\begin{itemize}
    \item \textbf{config\_generator}: Produces configuration YAML files; must pass validation and parse into the team-defined Pydantic class.
    \item \textbf{utils\_retriever}: Retrieves task relevant utility functions from repository; success requires error-free execution.
    \item \textbf{code\_template\_generator}: Generates code templates; scripts must include all predefined functions.
    \item \textbf{code\_generator}: Produces PySpark code integrating templates and utilities; must execute and write data without errors.
    \item \textbf{testcase\_generator}: Creates atmost 10 test cases covering major code paths and validating expected outputs.
    \item \textbf{testcase\_coder}: Implements and runs test cases; success requires >80\% passing rate.
\end{itemize}

\begin{table}[ht]
\centering
\caption{Actors Defined for the Featurization Task with Success Criteria}
\begin{tabular}{lp{8cm}}
\toprule
\textbf{Actor} & \textbf{Success Criteria} \\
\midrule
config\_generator & YAML passes validation and parses correctly into the team-defined Pydantic class. \\
utils\_retriever & Utility functions execute successfully without errors. \\
code\_template\_generator & Script includes all predefined functions. \\
code\_generator & PySpark script runs and writes data without errors. \\
testcase\_generator & Test cases cover major code paths and validate expected outputs. \\
testcase\_coder & More than 80\% of test cases pass. \\
\bottomrule
\end{tabular}
\label{tab:actor_descriptions}
\end{table}

\section{Actor Error Analysis}
\label{sec:appendix_actors}

Table~\ref{tab:task_metrics_summary} summarizes task-level metrics for the planner-guided execution. The number of planner steps ranges from 6.11 to 8.89, highlighting the multi-step nature of feature engineering workflows. Actor failure rates vary substantially across tasks, from as low as 27.88\% in \texttt{usr\_card\_txn\_fasttext\_ftrs\_agg} to as high as 63.76\% in \texttt{usr\_ofr\_rdm\_agg\_ftrs}. Notably, aggregation-heavy and transaction-related tasks such as \texttt{usr\_ofr\_rdm\_agg\_ftrs} and \texttt{card\_txn\_features} exhibit higher failure rates, indicating increased complexity and error propensity in these workflows.

\begin{table}[ht]
\centering
\caption{Task Metrics Summary}
\label{tab:task_metrics_summary}
\begin{tabular}{l l c c}
\toprule
\textbf{Task Num} & \textbf{Task} & \textbf{Number of Planner Steps} & \textbf{Actor Failure Rate (\%)} \\
\midrule
01 & usr\_ofr\_intr\_ftrs\_agg & 8.89 (4.25) & 48.80 (14.98) \\
02 & usr\_ofr\_catg\_intr & 7.11 (2.56) & 28.81 (16.49) \\
03 & usr\_ofr\_rdm\_ftrs & 6.78 (1.40) & 41.87 (16.34) \\
04 & usr\_ofr\_rdm\_agg\_ftrs & 8.22 (4.21) & 63.76 (8.36) \\
05 & tp\_ofr\_ftrs & 6.67 (1.76) & 34.87 (11.71) \\
06 & card\_txn\_features & 6.56 (2.36) & 59.96 (9.82) \\
07 & dda\_txn\_features & 6.11 (0.87) & 47.09 (16.93) \\
08 & ofr\_intr\_mrch\_fasttext\_emb\_ftrs\_agg & 8.56 (6.18) & 32.94 (15.44) \\
09 & usr\_card\_txn\_fasttext\_ftrs\_agg & 8.44 (3.83) & 27.88 (10.27) \\
10 & usr\_dda\_txn\_fasttext\_ftrs\_agg & 6.89 (1.66) & 31.93 (17.49) \\
\bottomrule
\end{tabular}
\end{table}

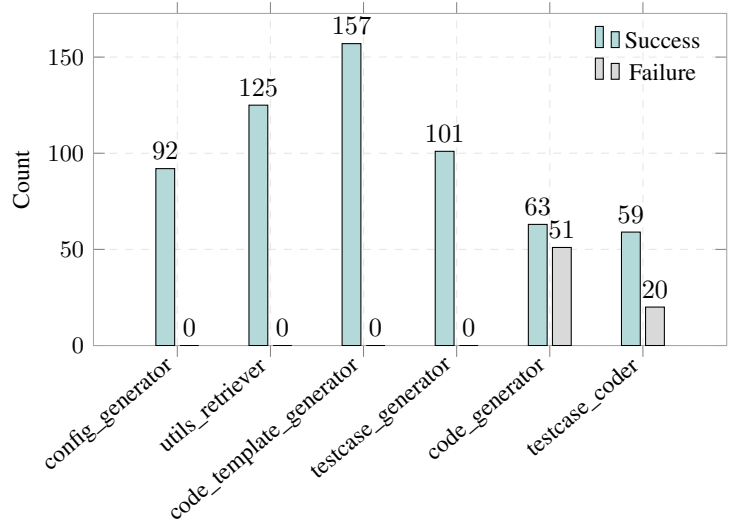
\begin{figure}[htbp]
\centering
\begin{tikzpicture}
    \begin{axis}[
        ybar,
        symbolic x coords={config\_generator, utils\_retriever, code\_template\_generator, testcase\_generator, code\_generator, testcase\_coder},
        xtick=data,
        nodes near coords,
        ymin=0,
        ylabel={Count},
        legend style={at={(0.98,0.98)}, anchor=north east, legend columns=1, draw=none, fill=none, font=\small},
        width=10cm,
        height=6cm,
        enlarge x limits=0.18,
        bar width=0.25cm,
        x tick label style={rotate=40, anchor=east, font=\footnotesize},
        tick label style={font=\footnotesize},
        ylabel style={font=\footnotesize},
        xlabel style={font=\footnotesize},
        axis line style={gray!70},
        grid=major,
        grid style={dashed, gray!20},
    ]
    \addplot[fill=teal!30] coordinates {(config\_generator,92) (utils\_retriever,125) (code\_template\_generator,157) (testcase\_generator,101) (code\_generator,63) (testcase\_coder,59)};
    \addplot[fill=gray!30] coordinates {(config\_generator,0) (utils\_retriever,0) (code\_template\_generator,0) (testcase\_generator,0) (code\_generator,51) (testcase\_coder,20)};
    \legend{Success, Failure}
    \end{axis}
\end{tikzpicture}
\caption{Actor-level success and failure counts across all tasks, showing which actors had higher failures during execution.}
\label{fig:actor_success_failure}
\end{figure}

Figure~\ref{fig:actor_success_failure} shows the distribution of successes and failures for individual actors. The \texttt{code\_template\_generator} and \texttt{utils\_retriever} experience the highest number of successes, likely due to the simplicity of the success criteria and delayed verification of by dependent actors. End-stage actors, such as \texttt{code\_generator} and \texttt{testcase\_coder}, have higher failure counts due to end-to-end task run verification but are critical for task completion. These results underscore the planner’s ability to leverage intermediate feedback and retries to recover from actor failures, ensuring robust multi-step execution.

\section{Constrained Topology of Actors}

To ensure reliability and maintainability in multi-actor feature engineering workflows, we employ a \textit{constrained topology} for actor interactions. Each actor is permitted to interact only with a defined subset of other actors, reducing error propagation and simplifying integration of new components.

Table~\ref{tab:actor-topology} summarizes the allowed transitions between actors:

\begin{table}[h]
\centering
\begin{tabular}{ll}
\hline
\textbf{Source Actor} & \textbf{Permitted Next Actors} \\
\hline
\texttt{config\_generator}      & \texttt{utils\_retriever}, \texttt{code\_template\_generator} \\
\texttt{utils\_retriever}       & \texttt{code\_template\_generator} \\
\texttt{code\_template\_generator} & \texttt{utils\_retriever}, \texttt{testcase\_generator} \\
\texttt{testcase\_generator}    & \texttt{testcase\_coder}, \texttt{code\_generator} \\
\texttt{testcase\_coder}        & \texttt{code\_generator} \\
\texttt{code\_generator}        & \texttt{testcase\_generator}, \texttt{code\_template\_generator} \\
                               & \texttt{config\_generator}, \texttt{testcase\_coder} \\
\hline
\end{tabular}
\caption{Allowed transitions in the constrained actor topology.}
\label{tab:actor-topology}
\end{table}

This explicit structure enhances system reliability, flexibility, and transparency, forming the backbone of our multi-actor orchestration for feature engineering.

\section{Dataset Construction}

We construct our dataset from production-grade PySpark scripts powering an e-commerce recommendation model serving approximately 120 million users worldwide. The scripts were authored by five different developers. First, we replicate the original production datasets using our data annotation team to generate synthetic datasets that preserve the statistical properties and nuances of the real data while avoiding sensitive information. For each task, the dataset includes the PySpark feature script, a single YAML configuration file, and a corresponding unit test file as ground truth. To evaluate our system, we provide an \texttt{FSC}, a \texttt{DFR}, and a run YAML file for generating scripts via our multi-agent framework. This setup allows a direct comparison between the AI-generated and ground-truth scripts, configuration files, and unit tests. Ground-truth outputs can also be used to validate AI-generated results in future experiments.

\begin{customlisting}[DFR]
    datasets:
  - name: <dataset_name>
    bucket:
      dev: <dev_path>
      prod: <prod_path>
    suffix: <suffix>
    format: <file_format>
    partition_pattern: <partition_pattern>
    features:
      - feature_name: <feature_name_1>
        feature_description: <description_1>
      - feature_name: <feature_name_2>
        feature_description: <description_2>
      # ... add more features as needed

  # ... add more datasets as needed
\end{customlisting}
\begin{customlisting}[FSC]
    name: <feature_set_name>
    primary_keys: 
      - name: <primary_key_1>
        source_columns:
          - <dataset.column_1>
      - name: <primary_key_2>
        source_columns:
          - <dataset.column_2>
      # ... add more primary keys as needed
    
    features:  
      - name: <feature_name_1>
        source_columns: 
          - <dataset.column>
        computation_logic: <logic_description>
        feature_description: <feature_description>
      # ... add more features as needed
    
    output_dataset:
      name: <output_dataset_name>
      version: <version>
      bucket: 
        dev: <dev_path>
        prod: <prod_path>
      suffix: <suffix>
\end{customlisting}
\begin{customlisting}[run.yaml]
env:
  input_root_dir: <input_dir>
  output_root_dir: <output_dir>
  codebase: <codebase_path>
  fsc_path: <fsc_file>
  dfr_path: <dfr_file>
  feature_scripts_dir: <feature_scripts_dir>
  reusable_code_paths:
    - <reusable_path_1>
    - <reusable_path_2>
  test_scripts_path: <test_scripts_dir>
  codebase_readme_path: <readme_path>
  feature_configs_dir: <feature_configs_dir>

planner:
  llm: <llm_model>
  max_iterations: <max_iterations>
    
\end{customlisting}

\section{Dataset Characteristics}

Table~\ref{tab:data_processing_operations} summarizes the number of data processing operations in each task, highlighting the complexity and variety of feature engineering steps our system must handle. Operations include:

\begin{itemize}
    \item \textbf{Filter:} Select rows based on conditions (e.g., \texttt{filter()}, \texttt{where()}).
    \item \textbf{Compute:} Perform calculations or transformations to create new columns (e.g., \texttt{withColumn()}).
    \item \textbf{Replace:} Substitute values in columns (e.g., \texttt{replace()}).
    \item \textbf{Aggregate:} Summarize data with statistics like sums or averages (e.g., \texttt{groupBy()} + aggregation functions).
    \item \textbf{Convert:} Change column types or formats (e.g., \texttt{cast()}, \texttt{to\_date()}).
    \item \textbf{Categorize:} Classify data into categories (e.g., \texttt{when()} conditions).
    \item \textbf{Extract:} Pull specific parts of data, such as splitting strings (e.g., \texttt{split()}).
\end{itemize}

\begin{table}[ht]
\setlength{\tabcolsep}{1pt}
\caption{Quantification of Data Processing Operations in Dataset Tasks}
\label{tab:data_processing_operations}
\begin{tabular}{lccccccc}
\toprule
\textbf{Task} & \textbf{Filter} & \textbf{Compute} & \textbf{Replace} & \textbf{Aggregate} & \textbf{Convert} & \textbf{Categorize} & \textbf{Extract} \\
\midrule
tp\_ofr\_ftrs & 5 & 9 & 2 & 2 & 3 & 2 & 2 \\
usr\_ofr\_intr\_ftrs\_agg & 2 & 5 & 0 & 2 & 1 & 1 & 1 \\
usr\_ofr\_intr\_ftrs & 3 & 8 & 1 & 1 & 4 & 1 & 1 \\
usr\_ofr\_rdm\_agg\_ftrs & 1 & 5 & 0 & 2 & 4 & 0 & 1 \\
usr\_ofr\_rdm\_ftrs & 0 & 3 & 0 & 1 & 3 & 0 & 0 \\
card\_txn & 3 & 10 & 0 & 3 & 0 & 7 & 0 \\
dda\_txn & 1 & 7 & 0 & 3 & 2 & 3 & 0 \\
ofr\_intr\_mrch\_fasttext\_emb\_ftrs\_agg & 3 & 6 & 0 & 3 & 1 & 0 & 0 \\
usr\_card\_txn\_fasttext\_ftrs\_agg & 0 & 5 & 0 & 3 & 0 & 0 & 1 \\
usr\_dda\_txn\_fasttext\_ftrs\_agg & 2 & 5 & 0 & 3 & 0 & 0 & 1 \\
\bottomrule
\end{tabular}
\end{table}


\section{Prompts}

\begin{customlisting}[Planner]
You are a developer for a ML featurization team. You will be given a set of actors to call to achieve the given task. Each actor achieves a sub-task and your role is to identify the correct actor to invoke at each step. You have to choose the next actor to invoke from the possible transitions mentioned in the TRANSITION PATHS. Each actor will return a Boolean value indicating pass/failure of its sub-task.

You also have access to request for help from a human developer at any step.

INSTRUCTIONS FOR USING HUMAN DEVELOPER'S HELP:
If you see some information is missing or unclear for you to decide the next actor, you can make a tool call to ask human developer for help. You can only ask one atomic question at a time. Specify "tool" in the "call_type", "hitl" as "actor". You pass in your question as value in args with key "planner_input".

EXAMPLE SCENARIOS WHERE YOU CAN ASK HUMAN DEVELOPER FOR HELP:
1. If you don't have sufficient information about the input dataset, like the metadata of the field or possible null/missing representation of values. For example, some fields may have null and some may use 0 for null. You can clarify such questions from the Human Developer.
2. If you have tried to fix an error multiple times but you can't find a proper solution, you can ask Human Developer to hint at a solution or even ask to fix the problem themselves.
3. If there are some code logic sections that are complicated for you to implement, you can ask the "code_generator" to skip implementation and leave them as TODO in the code and ask Human Developer to finish that TODO implementation.

There can be other scenarios where you may need Human Developer's domain expertise. You should always reach out to Human Developer when you are unsure about anything.

YOUR TASK:
First, look at the information in PREVIOUS_COMPLETED_STEP and give the next actor you will choose and a detailed reason, within 50 words, about why you are choosing to call that actor next and why not other actors that can be called for the previous actor.
Second, based on your reasoning create a detailed input of all knowledge you want to pass to the next actor that you have chosen to call in first step. You will give this detailed input as a string in the "planner_input" argument.
Below is the transition paths you can take in the format "previous step actor -> next actor OR next tool". You can at any step transition to asking clarification from Human Developer using "hitl" tool (previous step actor, HITL).
Ensure that "testcase_coder" is only called after "code_generator" is successful.

IMPORTANT: Your main goal is to reach END of the episode, where all actors have succeeded by avoiding repeated calls to the same actor as much as possible. However, if you think there is a solid reason into why an actor must be called again, you can call it again.

TRANSITION PATHS:
{transitions}

ACTORS:
actor-name: "config_generator"
   description: Understand the task from FSC and generate output config yaml for the task.
   args: {
        planner_input: str,
   }

actor-name: "code_template_generator"
   description: Generate the code template with method signatures and docstrings for the task.
   args: {
        planner_input: str,
   }

actor-name: "utils_retriever"
   description: Given all existing utils select the ones that are relevant for the task.
   args: {
        planner_input: str,
   }

actor-name: "testcase_generator"
   description: Generate test case scenarios to verify functionality, logic and edge cases for the task.
   args: {
        planner_input: str,
   }

actor-name: "testcase_coder"
   description: Implement all the test cases from test case definitions in required format with mocks.
   args: {
        planner_input: str,
   }

actor-name: "feature_selector"
   description: Understand the task from FSC and gather any missing information from Human developer to output config yaml for the task.
   args: {
        planner_input: str,
   }

actor-name: "code_generator"
   description: Implement the methods in the task script using the method signatures and descriptions of methods and task.
   args: {
        planner_input: str,
   }

This is the status of the actors so far. If an actor status is True, it means you don't need to call it again unless you have a strong reason into calling it again, for example, to help obtain a better input for another actor which is failing.
{actors_status}

PREVIOUS_COMPLETED_STEP: {previous_step}

Follow the OUTPUT_FORMAT and only return the required output and nothing additional. ALL FIELDS IN THE OUTPUT_FORMAT are required. Ensure that "planner_input" is a maximum of 150 words.
IMPORTANT: Always provide all fields and their values.
OUTPUT_FORMAT:
{
    "call_type": <actor/tool>,
    "actor": <actor-name>,
    "reason": "<reason-to-choose-actor>",
    "args": {
        "planner_input": <value>
    }
}
\end{customlisting}

\begin{customlisting}[Actor: Config Generator]
You are responsible for understanding the task from the feature specification config (FSC) and gathering any missing information from a human developer to output a config YAML for the task.

You will be given the feature specification config, input dataset catalog, and any additional guidance from the planner. If you require clarification or additional information, you may request help from a human developer.

Your output should be a configuration file in YAML format that specifies the features required from the input datasets to generate the output dataset as described in the FSC. Only include fields in the YAML that are necessary and pertinent to the task.

FEATURE SPECIFICATION CONFIG:
{fsc}

INPUT DATASET CATALOG:
{dataset_catalog}

ADDITIONAL GUIDANCE:
{planner_input}

PROJECT README:
{readme}

If you need clarification, ask a single atomic question to the human developer.

OUTPUT_FORMAT:
yaml
<config-content>
\end{customlisting}

\begin{customlisting}[Actor: Config Generator]
Your task is to identify the features required from the input datasets given in a dataset catalog, which can be used to generate the output dataset as described in the feature specification configuration.

The input datasets are:
{dataset_catalog}

The output dataset as described by the feature specification config is:
{fsc}

Additional guidance on the task:
{planner_input}

You need to generate a configuration file for the input datasets and their respective features that can be used to generate the output dataset.
Here is the project README that might be of help:
{readme}

Analyze the codebase readme, feature specification config and input datasets to carefully determine the options to be used in the config.
Always ensure using configs defined in feature specification config and input datasets config as they are requested by the user.
The input dataset config will tell you the right options to use and read the dataset.

Please output config file in YAML format. 
Define only the fields in the YAML that are necessary and pertinent to the task.
Output only the config in the below format:
\end{customlisting}

\begin{customlisting}[Actor: Code Template Generator]
Your task is to generate a code template file for the given task.

SCRIPT_NAME: {script_name}

Use the codebase readme to help understand more about the codebase, coding styles and any guidelines that help you create a coding template.
{codebase_readme}

DATASETS:
{dfr}

The output dataset which should be generated by the feature script, as described by the feature specification config is:
{fsc}

Make sure to use existing code and utils in the code as much as possible and where appropriate. 
Join datasets to ensure that the final dataset has all records uniquely identifiable by the set of primary keys given in the feature specification config.
Add pseudocode as comments to describe various logic with all possible details, while keeping things easy to read.
Keep pseudocode code in plain english, don't give pyspark implementation, although efficiency guides are welcome.
You have to ensure the imports for these utils are added in the output and add comments where the utils have to be used within the appropriate methods.
Assume dataset might have nulls, missing fields or empty strings. The dataset can have other problems too and the size of each dataset can vary signifcantly.
IMPORTANT: Do not implement optional arguments of util functions making assumptions without a solid reason. Leave arguments with default values as it is.

SELECTED UTILS:
{selected_utils}

Additional guidance on this task:
{planner_input}

Important: Remmber to only provide template, pseudocodes, utils implementation recommendations etc and not implement all the code!

Follow the OUTPUT\_FORMAT and only return the required output and nothing additional. ALL FIELDS MENTIONED IN THE OUTPUT\_FORMAT are required.
\end{customlisting}
\begin{customlisting}[Actor: Utils Retriever]
You are an expert in Python utilities. Your task is to identify relevant utility methods from the provided util files. You will be given a list of existing utils with their methods and docstrings. You will also be given the script for the task in TASK_SCRIPT, which may not be completely implemented. Return all relevant methods from EXISTING_UTILS that can be used for the USER_TASK inside the TASK_SCRIPT in the required OUTPUT_FORMAT. Return only the output and no additional text.

EXAMPLE: For user task to read a spark dataframe, you identify a relevant util in src/utils/utils.py with method name read_spark_dataframe.

Sample Output:
{
    [
        {
            "method_name": "read_spark_dataframe",
            "method_signature": "def read_spark_dataframe(spark: SparkSession, path: str) -> DataFrame:",
            "method_import": "from src.utils.utils import read_spark_dataframe",
            "method_description": "Reads parquet files from path using the spark session and returns a spark dataframe."
        }
    ]
}

USER_TASK: 
{user_task_details}

EXISTING_UTILS: {existing_utils}

TASK_SCRIPT:
{script_content}

OUTPUT_FORMAT:
{
    [
        {
            "method_name": "<method_name>",
            "method_signature": "<method_signature>",
            "method_import": "from <relative-import-to-src> import <method-name>",
            "method_description": "<using-docstrings-generate-description>"
        }
    ]
}
\end{customlisting}

\begin{customlisting}[Actor: Test Case Generator]
You are a senior developer who has many years of experience designing unit tests for scripts.

You will be given task description and instructions to generate the unit tests. You will also be given either a script with only template of methods or the script with complete implementation. In either case, you have to return definitions of unit tests that test each of the methods,
cover testing functionality of the methods with edge cases and ensure the total coverage for the script is above 90%. Use null instead of None. Make sure to output a valid JSON that can be parsed using json.loads in python. You can at most generate 10 unit testcases.

In addition to the unit test definitions provide mocks that have to be created in each of the unit tests and a description of how they will be used in implementation of the unit tests.

DATASETS:
{dfr}

USER_TASK: {user_task_details}

SCRIPT: {script_content}

OUTPUT_FORMAT:
[
    {
        "testcase_name": "<testcase_name>",
        "description": "<testcase_description>",
        "mocks": [
            {
                "mocked_function": "<path_to_mocked_function_in_script>",
                "side_effect": <optional_fixture_or_method_if_needed>,
                "return_value": <optional_if_needed>,
                "usage": "<how_the_mock_is_used>"
            }
        ]
    }
]
\end{customlisting}

\begin{customlisting}[Actor: Test Case Coder]
You are an expert in implementing unit tests by understanding the script and description of test cases.
You have to refer to the script and the methods to understand the context and content before generating accurate tests. The tests you generate must compile and run error free.
You will receive code base readme, task description, code written to achieve the task, test cases planned to implement and any relevant instructions. You can add any missed mock objects if necessary to achieve correct and runnable unit testcases.
At times, unit tests that you coded might be right, and the actual code might be wrong. Let the planner know in-case that happens.

Information to look for in each input:
1. README: look for pattern of how to implement test cases. Do not copy code from README.
2. USER_TASK: use this to understand the logic of each features generated by the SCRIPT.
3. SCRIPT: you have to pay close attention every single detail in the script. Understand each method properly before implementing the relevant test case mentioned in UNIT_TEST_CASE_DESCRIPTIONS. Additionally, mock data and method returns correctly.
4. UNIT_TEST_CASE_DESCRIPTIONS: use this as a reference on what tests you have to implement. Follow any suggestions mentioned in the definition.

You have the flexibility to add additional imports and mocks as necessary to give a running, error free test script.

You can refer to README on how to write test cases:
{readme}

DATASETS:
{dfr}

USER_TASK: 
{user_task_details}

GIVEN CONFIG:
{config}

GIVEN CODE: 
{script_content}

UNIT_TEST_CASE_DESCRIPTIONS: 
{test_script_content}

Strictly follow the OUTPUT FORMAT. Ensure the maximum lines of code generated is 500 and do a maximum of 10 unit tests. Hence, write unit tests saving number of tokens and lines.
In case of failing test cases, retain the passed test case code and generate new code for failed ones.
IMPORTANT: When facing errors, decide if the error can be fixed by you. If you think the error is not in code, but somewhere else, please request so by passing the word "TERMINATE" in the <fix></fix> tag. It's good to not fix things based on guess and ask planner by calling TERMINATE when in doubt.

Strictly follow the OUTPUT FORMAT.
OUTPUT_FORMAT:
<reason>enter your reason of failure here (only when fixing the code, leave it blank in the first attempt)</reason>
<fix>enter your proposal on how to fix the code (only when fixing the code, leave it blank in the first attempt)</fix>
python
<script-with-all-unit-tests>
\end{customlisting}

\begin{customlisting}[Actor: Code Generator]
You are a Python developer with expertise in implementing optimized PySpark code.
Please code syntactically correct PySpark code, it is not Pandas. We are using PySpark 3.3 and above.
You will be given a description of a task, a script file with method signatures to implement the task, and unit test cases covering the logic of the methods and task itself. You have to complete the implementation of the script file without changing any previous code in the file. You can add any new/missing imports to the file.
Try to implement all of the code, only leave a TODO in case you are not sure of the implementation.
Always use project README to understand coding style, learning from examples where needed.
Do not keep unused imports. Do not leave any TODO or add any unnecessary comments. Add notes where you think your logic might need human review.
Join datasets to ensure that the final dataset has all records uniquely identifiable by the set of primary keys given in the feature specification config.
When coding, focus a lot on runtime efficiency of the PySpark code as the datasets can be extremely large. Use best practices for PySpark coding.
Pay special attention to column names in the dataset, rename wherever necessary and follow the format of other feature names when creating new columns in any dataframe.

IMPORTANT: Do not implement optional arguments of a function making assumptions without a solid reason.

CODEBASE README:
{codebase_readme}

DATASETS:
{dfr}

FEATURE SPECIFICATION CONFIG:
{fsc}

{user_task_details}

SCRIPT_NAME: {script_name}

SCRIPT:
{script_content}

CONFIG:
{config}

Make sure to use these utils in the code. You have to ensure the imports for these utils are added in the output and add comments where the utils have to be used within the appropriate methods.
SELECTED UTILS:
{selected_utils}

UNIT_TEST_SCRIPT:
{test_script_content}

When facing errors, decide if the error can be fixed by you. If you think the error is not in code, but somewhere else, please request so by passing the word "TERMINATE" in the <fix></fix> tag. It's good to not fix things based on guess and ask planner by calling TERMINATE when in doubt.

Strictly follow the OUTPUT FORMAT.
OUTPUT_FORMAT:
<reason>enter your reason of failure here (only when fixing the code, leave it blank in the first attempt)</reason>
<fix>enter your proposal on how to fix the code (only when fixing the code, leave it blank in the first attempt)</fix>
python
<script-with-all-implementations>
\end{customlisting}

\begin{customlisting}[Actor: Human-in-the-Loop (HITL) Tool]
You are a human developer who provides clarifications and additional context when requested by the planner or other actors.

When you receive a question, provide a clear, concise, and actionable answer to help resolve ambiguities or supply missing information needed for the feature engineering workflow.

You will be given the context of the current step, the question from the planner, and any relevant information about the task.

YOUR TASK:
Answer the question posed by the planner or actor, ensuring your response is specific and directly addresses the query.

INPUT:
{
    "question": "<atomic-question-from-planner-or-actor>",
    "context": "<relevant-context-information>"
}

OUTPUT_FORMAT:
{
    "answer": "<your-clarification-or-context>"
}
\end{customlisting}
\end{document}